\documentclass[10pt,twocolumn,letterpaper]{article}

\usepackage{cvpr}              

\usepackage{graphicx}
\usepackage{amsmath}
\usepackage{amssymb}
\usepackage{booktabs}
\usepackage{algorithm}
\usepackage{algorithmic}
\usepackage[accsupp]{axessibility}
\usepackage[pagebackref,breaklinks,colorlinks]{hyperref}

\usepackage[capitalize]{cleveref}
\crefname{section}{Sec.}{Secs.}
\Crefname{section}{Section}{Sections}
\Crefname{table}{Table}{Tables}
\crefname{table}{Tab.}{Tabs.}

\def\cvprPaperID{*****} 
\def\confName{CVPR}
\def\confYear{2023}

\begin{document}

\title{Federated Learning in Non-IID Settings Aided by Differentially Private Synthetic Data}

\author{Huancheng Chen, Haris Vikalo\\
Universitiy of Texas at Austin, Austin, Texas\\
{\tt\small huanchengch@utexas.edu,\tt\small hvikalo@ece.utexas.edu}
}
\maketitle

\begin{abstract}
   Federated learning (FL) is a privacy-promoting framework that enables potentially large number of clients to collaboratively train machine learning models. In an FL system, a server coordinates the collaboration by collecting and aggregating clients' model updates while the clients' data remains local and private. A major challenge in federated learning arises when the local data is non-iid -- the setting in which performance of the learned global model may deteriorate significantly compared to the scenario where the data is identically distributed across the clients. In this paper we propose FedDPMS (\underline{Fed}erated \underline{D}ifferentially \underline{P}rivate \underline{M}eans \underline{S}haring), an FL algorithm in which clients augment local datasets with data synthesized using differentially private information collected and communicated by a trusted server. In particular, the server matches the pairs of clients having complementary local datasets and facilitates differentially-private sharing of the means of latent data representations; the clients then deploy variational auto-encoders to enrich their datasets and thus ameliorate the effects of non-iid data distribution. Our experiments on deep image classification tasks demonstrate that FedDPMS outperforms competing state-of-the-art FL methods specifically developed to address the challenge of federated learning on non-iid data.
\end{abstract}

\vspace{-0.2 in}
\section{Introduction}
\label{sec:introduction}

The need for massive amounts of high-quality training data in deep learning (e.g., ImageNet \cite{lecun1989backpropagation}, COCO \cite{lin2014microsoft}) creates a major challenge in settings where data is distributed across a potentially large number of users' devices; in particular, constraints on communication resources and users' privacy concerns often prohibit gathering local data and training models at a central location. In response, {\it federated learning} (FL) where users collaboratively train a global model without revealing personal data has emerged as a privacy-promoting and communication-efficient distributed alternative to centralized learning \cite{kairouz2019advances,yang2019federated, li2020federated,generative}.

In FL systems, a subset of users' devices updates a global model by training on local data; a server coordinates the training process by selecting the users, collecting the updates, and aggregating them to form a new global model. The original FL algorithm, {\it Federated Averaging} (FedAvg) \cite{mcmahan2017communication}, chooses users at random and updates the global model by averaging the users' updates;
the convergence analysis provided in \cite{mcmahan2017communication} assumes that local datasets are independent and identically distributed (i.i.d.). Recently, improving performance of FL systems that deploy sophisticated ML models in a variety of practical scenarios has received considerable amount of attention \cite{FedProx,scaffold,federatedunsupervised,adaptive,FedDT,fedboost,fedgan,pipetransformer}.

A major challenge in FL presents when the decentralized data is heterogeneous. Indeed, it is unrealistic to expect that the users participating in a FL system train on data generated from identical distributions -- instead, distributions of labels will likely differ between the participating devices. This is particularly pronounced in setting where the training data is limited, possibly to the extent that only a subset of classes is present in the users' datasets. Since in such scenarios performance of FedAvg may significantly deteriorate, a number of approaches for learning from heterogeneous distributed data has recently been proposed \cite{FedProx, scaffold,moon}. Data heterogeneity in FL systems deploying deep learning networks and training on complex datasets was studied in \cite{moon} where a regularization term is introduced in order to impose contrastive learning on local updates, effectively aligning those updates with the global objective. However, as we illustrate in our experiments (see Section \ref{exp}), this approach fails to perform well on imbalanced data partitions. 
Difficulties in learning a global model under data heterogeneity have also motivated various clustering and personalization approaches to FL including Model-Agnostic Meta-Learning (MAML) and its variants that rely on client clustering \cite{FVKL21, SZC+20, FAZY21,GCYR20,SMS20,XLS21}.

In this work we address the challenge of data heterogeneity and scarcity in learning a global model by enabling clients to locally synthesize data using parameters acquired and shared in a privacy-preserving manner. In particular, the proposed {\it federated differentially-private means sharing} (FedDPMS) framework allows each client to locally generate synthetic data by relying on means of latent data representations exchanged in differentially-private (DP) manner between clients matched by a trusted server. The server coordinating the training process is assumed to be provided partial information about the clients' data distributions (in particular, indices of the most abundant and the least abundant classes), and is therefore capable of matching the pairs of clients whose local datasets are complementary. Unlike the existing data augmentation approaches to FL which utilize Generative Adversarial Networks (GANs) \cite{GAN,federatedGAN,fedgenerative}, our proposed framework relies on simple Variational Auto-Encoders (VAEs) well-suited to the classification tasks of interest. Privacy of sharing the (noisy) means of latent representations is quantified by utilizing the differential privacy mechanism. To test the proposed FedDPMS framework, we conduct extensive numerical studies on image classification tasks using Fashion-MNIST \cite{xiao2017/online}, CIFAR-10 and CIFAR-100 datasets. In those experiments FedDPMS outperforms state-of-the-art approaches, particularly in settings where the local data comes from diverse distributions.

\vspace{-0.05 in}
\section{Related Work}
\vspace{-0.05 in}
\label{related work}
\subsection{Federated Learning on Heterogeneous Data}
\label{HeterogeneousFL}
\vspace{-0.05 in}


Approaches to FL that aim to overcome challenges of training a global model on heterogeneous data can be broadly organized in two categories. The first category includes techniques that attempt to improve the model aggregation step performed by the server. Examples include PFNM \cite{bayesian}, a Bayesian non-parametric approach for extracting layers from local models and using them to update the corresponding layers in the global model. While PFNM targets relatively simple architectures, FedMA \cite{fedMA} takes a step further and extends the same ideas to CNNs and LSTMs. In \cite{wang2020tackling}, the authors present a framework for the analysis of convergence of FL on heterogeneous data, along with a normalized averaging method, FedNova \cite{wang2020tackling}, that aims to eliminate objective inconsistencies (i.e., prevent convergence of the global model to a stationary point of the mismatched objectives) caused by naive aggregation of local models. The second category of methods for FL on heterogeneous distributed data is focused on reducing the drift in local training. To this end, FedProx \cite{FedProx} introduces a proximal term to the learning objective of each client with the goal of making local training aligned with the global objective. 
SCAFFOLD \cite{scaffold} utilizes predictive variance reduction \cite{accelerating} to introduce control variates and correct local updates. The above two studies were verified in experiments on MNIST and EMINIST with multinomial logistic regression and fully connected two-layer networks. Recently, \cite{moon} tested FedProx and SCAFFOLD on more challenging tasks that involve deep learning models, showing that those two methods unfortunately offer little to no advantage over FedAvg in the considered scenarios. As an alternative, \cite{moon} propose the Moon algorithm motivated by the local/global model proximity idea of FedProx but instead of the $l_2$-norm term, the proximity is imposed via a contrastive term in the objective of local training. 
Another study, FedDyn \cite{dynamicregular}, proposes a dynamic regularizer to promote convergence of the local loss to a stationary point of the global loss. Building on top of SCAFFOLD, FedDC \cite{feddc} introduces an auxiliary local drift variable which serves as a dynamic regularizer helping narrow the gap between local models and the global model. While the above approaches utilize different loss functions, ultimately they all deploy the same strategy of introducing a regularization term to prevent from overfitting in local training. 

\vspace{-0.05 in}
\subsection{Domain Generalization in Federated Learning}
\vspace{-0.05 in}
A set of methods complementary to the techniques in Section~\ref{HeterogeneousFL} relies on domain generalization, a method which aims to improve generalization ability of models by training them on data from multiple source domains. Such techniques include FedDG \cite{feddg}, a method that allows each client to transfer its amplitude spectrum, decomposed from raw data, to a bank at the server. The server shares the collected amplitude spectrum with all clients, enabling them to synthesize new distribution via interpolation and helping improve local training. However, FedDG is unable to enrich class diversity since the phase spectrum is not shared due to privacy concerns. Another study, \cite{hao2021towards}, employs zero-shot data augmentation and relies on the statistics of the batch normalization (BN) layers to reduce the variance of test accuracy. FedMix \cite{fedmix} presents a framework for sharing clients' averaged local data via Mixup \cite{mixup}; privacy of clients in FedMix may still be compromised since the only effort to protect it is based on averaging raw data. FedDA \cite{fedda} utilizes the attention mechanism \cite{vaswani2017attention} to enable the server to aggregate local per-label knowledge and create global per-label knowledge, ultimately enabling data augmentation via conditional variational autoencoder \cite{cvae}. However, aggregation of per-label knowledge in highly non-iid setting (e.g., only a few classes are present in the local dataset) presents the same challenge: local per-label knowledge is drifting from the expectation of the global per-label knowledge. FedDPMS aims to address this problem by matching pairs of client and generating complementary data using shared mean of latent representations.
\vspace{-0.05 in}
\section{Differentially Private Mean Sharing}
\vspace{-0.05 in}
\subsection{Overview of the Proposed Scheme}
\vspace{-0.05 in}
Both existing approaches to dealing with data heterogeneity  -- improving model aggregation at the server and reducing model drift in local training -- struggle when the differences between local data distributions are significant (e.g., the clients completely missing some class labels). To this end, in this paper we propose an alternative framework that aims to provide privacy-preserving domain generalization by enhancing and balancing local datasets. In particular, FedDPMS relies on sharing differentially-private information needed to generate representative synthetic data -- specifically, the clients share noisy versions of averaged latent representations that their encoders extract from local data. Since the proposed framework facilitates learning in non-iid settings by enriching the diversity of clients' data, it does not compete with the prior work in Section~\ref{related work} -- instead, FedDPMS is complementary to and may potentially be combined with the existing methods for FL on non-iid distributed data. In the upcoming discussions, for illustration purposes we will repeatedly invoke the image classification task as a use case.
\vspace{-0.05 in}
\subsection{Network Architecture and Training Objective}
\vspace{-0.05 in}
\label{trainingphases}
The global model and local models in an FL system typically share the same network structure. In our proposed framework, the model consists of three components: an encoder, a decoder and a classifier. The encoder learns data representation utilizing convolutional layers; in particular, data samples are encoded into a mean $\mu$ and a variance $\sigma_{z}^2$, and represented by a latent variable $z \sim {\cal N}(\mu,\sigma_{z}^2)$. Given the encoder's architecture, a symmetric decoder is designed by relying on up-sampling and transposed convolutions. The final component of the model is a classifier that consists of multiple linear layers and maps the latent variable $z$ into a categorical vector $p$ which quantifies the likelihood of each class. The described architecture and the principle of Variational Auto-Encoder (VAE) are illustrated in supplementay material section 1 and 6.

The training starts with a preliminary step which is focused on minimizing a loss that consists of three components: $\ell_1$, the cross-entropy between a prediction and the ground truth; $\ell_2$, the Kullback-Leibler divergence between the prior distribution $p(z)$ and the approximated distribution $q(z)$ (parametrized by mean and variance in the Gaussian case); and $\ell_3$, the reconstruction loss defined as the mean-square error between the original and reconstructed data points. Formally,
\[\ell_{1} = \operatorname{CEloss}\left(F_{w_{i}^{t}}(x), y\right), \; \ell_{2} = \operatorname{KLDloss}\left(\mu, \sigma_{z} \right),\]
\[\ell_{3} = \operatorname{MSEloss}\left(\hat{x}, x\right)\]
where $F_{w_{i}^{t}}(\cdot)$ denotes the entire network with parameters $w_{i}^{t}$, $x$ and $\hat{x}$ are the input and its reconstruction, respectively, and $y$ denotes the ground-truth. The local training loss is formed as
\[\ell = \ell_1 + \lambda(\ell_2 +\ell_3),\]
where $\lambda$ is a hyper-parameter. In our experiments, we fine-tune the hyper-parameter and report the performance with the best $\lambda$. The training then enters the secondary phase, where we train only the encoder and classifier (i.e., the optimization focuses on $\ell_1$). This is elaborated next.
\vspace{-0.05 in}
\subsection{The Proposed Learning Scheme -- FedDPMS}
\vspace{-0.05 in}
As stated in Section~\ref{trainingphases}, the training of FedDPMS is organized in two stages. I. {\it Preliminary training.} Each client locally trains a VAE and sends the parameters of the resulting encoder and classifier to the server; the server aggregates the received information and shares the resulting global model (encoder and classifier) with the clients (except in the last round of training, please see Section~\ref{Preliminary}). II. {\it Secondary training.} There are four steps in each round $t$ of the secondary training. (1) \textbf{Client matching}: The server is given partial information about the selected clients' data distributions (i.e., indices of the most and the least abundant classes) but has no access to raw data which remains private; based on the received information, for each client the server identifies the most informative latent representation statistics (specifically, the most informative means of latent representations); (2) \textbf{Data synthesis}: The server communicates (noisy) means of latent representations and the global decoder to the clients in need of certain data classes; using the global decoder and the information received from the server, these clients synthesize samples of locally missing or underrepresented classes and incorporate the synthesized samples into their local datasets. (3) \textbf{Model training}: the selected clients locally train encoders and classifiers using augmented datasets; the updated models are collected and aggregated by the server. (4) \textbf{DPMS}: The selected clients who prior to the current training round have not shared latent representation information upload the indices of their top-$n$ most abundant classes and differentially-private means of latent data representation to the server. The training procedure outlined in this section is formalized as Algorithm~\ref{alg:Framwork}. In the next two subsections, we provide further informative details of the preliminary and secondary training.
\begin{algorithm}[htb]

\caption{(FedDPMS)}
\label{alg:Framwork}
\begin{algorithmic}[1] 
\REQUIRE ~~\\ %
    Local datasets from $K$ clients, $\mathcal{D} = \{\mathcal{D}_{1},\mathcal{D}_{2},\dots\mathcal{D}_{K}\}$; the number of participating clients $k$ each round; the number of preliminary training rounds $T_p$;
    the number of global epoches $T$.
\ENSURE ~~\\ 
    The final global model $w^{T}$
    \STATE \textbf{Server executes: } 
    \STATE randomly initialize $(w_{e}^{0},w_{c}^{0})$, $\mathcal{A} = \emptyset, \mathcal{B} = \emptyset,  \mathcal{C} = \emptyset,\mathcal{Z} = \emptyset$
    \FOR{$t = 0,\dots,T_{p}-1$}
    \STATE $\mathcal{S}_{t}  \xleftarrow{} k$ clients selected at random
    \STATE $ (w_{e}^{t+1},w_{c}^{t+1})  \xleftarrow{} \textbf{PreTrain}(t, w_{e}^{t}, w_{c}^{t},\mathcal{S}_{t})$
    \ENDFOR

    \FOR{$t = T_{p},\dots,T-1$}
        \STATE $\mathcal{S}_{t}  \xleftarrow{} k$  clients selected at random
        \STATE $ (w_{e}^{t+1},w_{c}^{t+1})  \xleftarrow{}$
        \STATE $\textbf{SecTrain}(w_{e}^{t}, w_{c}^{t},\mathcal{S}_{t},\mathcal{A},\mathcal{B},\mathcal{C},\mathcal{Z})$
    \ENDFOR

\STATE \textbf{return} $w^{T} = (w_{e}^{T},w_{c}^{T})$ \\
\end{algorithmic}
\end{algorithm}
\vspace{-0.1 in}
\subsubsection{Preliminary training}
\vspace{-0.05 in}
\label{Preliminary}
During preliminary training, each client trains a VAE capable of compressing/reconstructing data and recognizing data labels. To curtail communication costs, only the encoder and classifier's weights are uploaded to the server (except in the last round of the preliminary training). Formally, at the beginning of global round $t$, the server sends the global encoder and classifier models $\{w_{e}^{t},w_{c}^{t}\}$ to the clients. The clients use $\{w_{e}^{t},w_{c}^{t}\}$ to initialize training VAEs on local data (decoder is not initialized), and run several epochs of the stochastic gradient descent; by the end of those epochs, the $i^{th}$ client has obtained updates  $\{w_{i,e}^{t},w_{i,c}^{t},w_{i,d}^{t}\}$. Finally, the server aggregates the updates $\{w_{i,e}^{t},w_{i,c}^{t}\}$ into the latest global model $\{w_{e}^{t+1},w_{c}^{t+1}\}$. In the last round of preliminary training, each selected client is requested to upload the full update $\{w_{i,e}^{t},w_{i,c}^{t},w_{i,d}^{t}\}$ so that the server can aggregate global decoder $\mathbf{w_{d}}$; at that point, local decoders may be deleted to free up memory.
\vspace{-0.15 in}
\subsubsection{Secondary training}
\vspace{-0.1 in}
\label{secondarytraining}
Recall that in this stage clients update the encoder and classifier by optimizing the cross-entropy between a prediction and the ground truth while disregarding the divergence and reconstruction loss; these updates are formed by training on augmented datasets. For the sake of computation and communication efficiency, each selected client shares its means of latent data representations and augments the local dataset only once, regardless of how many times the client is selected by the server; as our results demonstrate, this is sufficient to improve the performance of the learned model. To avoid repeating information sharing / dataset augmentation, the server maintains four sets: (1) the set of clients who shared information (i.e. who encoded original data into latent representations and shared noisy means of latent representations to the server), $\mathcal{A}$; (2) the set of clients who benefited from the shared information (i.e., who augmented their local datasets with synthesized artificial data), $\mathcal{B}$; (3) the set $\mathcal{C} = \{\mathcal{C}_{i}| i \in \mathcal{A}\}$, where $\mathcal{C}_{i} = (\mathcal{C}_{i}^{1},\dots,\mathcal{C}_{i}^{n})$ indicates the $n$ most abundant classes in the local dataset of client $i \in \mathcal{A}\}$; and (4) the set of shared latent representation means, $\mathcal{Z}= \{\mathcal{Z}_{i} | i\in \mathcal{A}\}$, where $\mathcal{Z}_{i}=\{(\tilde{z}_{i,1}^{1},\dots,\tilde{z}_{i,\alpha}^{1}),\dots, (\tilde{z}_{i,1}^{n},\dots,\tilde{z}_{i,\alpha}^{n})\}$, $\alpha$ is the number of repeatedly generated noisy means $\tilde{z}_{i}^{c}$ of the latent representation of the (abundant) class $c$ in the local dataset available to client $i$. These four sets are initialized as empty at the beginning of the secondary training. The described procedure for secondary training is formalized as Algorithm~\ref{alg:secondary}.
\label{app_alg2}
\begin{algorithm}[htb]
\caption{SecTrain}
\label{alg:secondary}
\begin{algorithmic}[1] 
\REQUIRE ~~\\ %
    global model $w_{e}^{t}, w_{c}^{t}$; selected $k$ clients $\mathcal{S}_{t}$; assisting clients $\mathcal{A}$; benefited clients $\mathcal{B}$; abundant class $\mathcal{C}$; shared means $\mathcal{Z}$; local datasets at $k$ clients, $\mathcal{D} = \{\mathcal{D}_{i}| i\in \mathcal{S}_{t}\}$; generation quota $\alpha$; standard deviation of additive noise $\sigma$.

\ENSURE ~~\\ 
    The global model $w_{e}^{t+1}, w_{c}^{t+1}$
    \STATE $\mathcal{R} \xleftarrow{} \textbf{Matching}(\mathcal{S}_{t}, \mathcal{A}, \mathcal{C}) $
    \FOR{$i \in \mathcal{S}_{t}$ \textbf{in parallel}}
        \IF{$\mathcal{R} \not = \emptyset$ \textbf{AND} $i \notin \mathcal{B}$}
        \IF{client $i$ did not download $\mathbf{w_{d}}$}
        \STATE download the global decoder $\mathbf{w_{d}}$ 
        \ENDIF
        \STATE download $\mathcal{Z}_{\mathbf{R}_{i}}$ from the server
        \STATE $\tilde{\mathcal{D}}_{i} \xleftarrow{} \textbf{Synthesis}(\mathbf{w_{d}},\mathcal{Z}_{\mathbf{R}_{i}}$)
        \STATE $\bar{\mathcal{D}}_{i} \xleftarrow{}$ $\tilde{\mathcal{D}}_{i}\cup \mathcal{D}_{i}$
        \STATE the server executes: $\mathcal{B} \xleftarrow{} \mathcal{B} \cup i$
        \ENDIF
        \STATE download $w_{e}^{t},w_{c}^{t}$ from the server
        \STATE upload: $(w_{e,i}^{t},w_{c,i}^{t})  \xleftarrow{} \textbf{Optim}(w_{e}^{t}, w_{c}^{t},\bar{\mathcal{D}}_{i})$
        \IF{$i \notin \mathcal{A}$}
        \IF{client $i$ did not download $\mathbf{w_{d}}$}
        \STATE download the global decoder $\mathbf{w_{d}}$ 
        \ENDIF
        \STATE  $(\mathcal{C}_{i}, \mathcal{Z}_{i}) \xleftarrow{} \textbf{DPMS}(w_{e,i}^{t},w_{c,i}^{t},\mathbf{w_{d}}, \mathcal{D}_{i}, \alpha,\sigma)$
        \ENDIF
        
    \ENDFOR
    \STATE \textbf{Server executes:}
    \STATE $(\mathcal{A},\mathcal{C}, \mathcal{Z})\xleftarrow{}(\mathcal{A}, \mathcal{C}, \mathcal{Z}) \cup_{i\in \mathcal{S}_{t}} (i,\mathcal{C}_{i},\mathcal{Z}_{i})$
    \STATE $|\bar{\mathcal{D}}^{t}| \xleftarrow{} \sum_{i \in \mathcal{S}_{t}} |\bar{\mathcal{D}}_{i}|$
    \STATE $w^{t+1}  \xleftarrow{} \sum_{i \in \mathcal{S}_{t}} \frac{|\bar{\mathcal{D}}_{i}|}{|\bar{\mathcal{D}}^{t}|}(w_{e,i}^{t}, w_{c,i}^{t})$
\STATE \textbf{return} $w^{t+1} $ \\
\end{algorithmic}
\end{algorithm}
\\
\textbf{Client matching.}
Let $\mathcal{S}_{t}$ denote the set of clients selected in training round $t$. Based on the information about the most and least abundant classes in local datasets, the server decides for each client who should they receive assistance from (i.e., which information in $\mathcal{Z}$ should they be given). The matching is based on the distance between the clients' data distributions; in particular, client $i$ is scheduled to be the recipients of client $j$'s latent space information if the server identifies that client $j$'s data distribution is such that the samples drawn from it would significantly diversify client $i$'s data.\footnote{For convenience, we refer to client $i$ as the ``benefiting client" and to client $j$ as the ``assisting client".} Specifically, each client $i \in \mathcal{S}_{t}$ still seeking to diversify local data sends indices of $n$ classes in its dataset with the fewest samples to the server; let $\mathcal{H}_{i}$ denote the set of the $i^{th}$ client's ``data scarce" classes. Having received $\mathcal{H}_{i}$, the server identifies client $j$ having the set of data abundant classes $\mathcal{C}_{j}$ that intersects with $\mathcal{H}_{i}$ more than any other set of data abundant classes. After matching, $\mathcal{Z}_{j}$ and 
the global decoder are sent to client $i$. Note that client $j$'s means of latent data representations $\mathcal{Z}_{j}$ are sent to client $i$ by the server -- there is no direct connection between clients $i$ and $j$. 

\begin{figure*}[t]
\begin{center}
\includegraphics[width= 1 \linewidth]{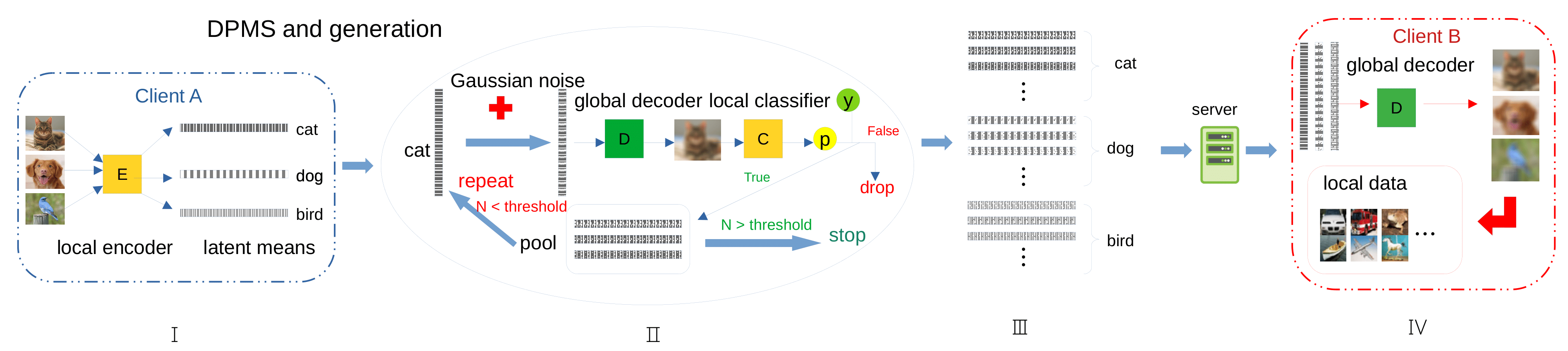}
\end{center}
   \caption{FedDPMS and synthetic data generation. The four parts of the figure depict: (1) finding latent representation of raw data via a local encoder; (2) creating noisy latent means (by adding Gaussian noise to the means of latent data representations) and filtering out unusable ones with the help of a local classifier; (3) uploading usable noisy latent means to the server; (4) a benefiting client utilizing the global decoder to generate synthetic data from the received noisy latent means, expanding its local dataset.}
\label{dpms}
\vspace{-0.2in}
\end{figure*}

In the first round of secondary training, the server does not pursue matching since $\mathcal{A}, \mathcal{B}, \mathcal{C}$ and $\mathcal{Z}$ are empty sets; the sets are augmented with new elements in the DPMS step. While our experiments demonstrate remarkable performance improvements of FedDPMS over competing methods despite providing a client in need of synthetic data with the latent representation of only one of its peers, the matching algorithm can readily be extended to identifying several ``assisting" clients for a ``benefitting"  client. The proposed method for client matching is formalized as Matching Algorithm in supplementary material section 3. \\
\textbf{Data synthesis.}
When client $i$ receives a set of noisy means of latent data representations and the global decoder, it utilizes the global decoder $\mathbf{w_{d}}$ to generate synthetic data $\tilde{\mathcal{D}_{i}}$; the synthetic data is merged with the local dataset $\mathcal{D}_{i}$ to form $\bar{\mathcal{D}_{i}}$.  The client then proceed to perform local model update by training on $\bar{\mathcal{D}_{i}}$, and uploads the result to the server. Upon receiving the update, the server adds index $i$ to the set of indices of clients who completed data diversification, $\mathcal{B}$. Samples of synthetic images are provided in supplementary material section 4.\\
\textbf{Model training.}
After generating synthetic data, each client $i$ has access to an augmented local dataset  $\bar{\mathcal{D}_{i}}$. When updating the model, the client samples $|\mathcal{D}_{i}|$ points uniformly at random from $\bar{\mathcal{D}_{i}}$ and utilizes only the sampled points for gradient computation ($|\cdot|$ denotes the set cardinality). This is to maintain the same complexity of the update step as FedAvg.
After generating synthetic data, each client $i$ has access to an augmented local dataset  $\bar{\mathcal{D}_{i}}$. When updating the model, the client samples $|\mathcal{D}_{i}|$ points uniformly at random from $\bar{\mathcal{D}_{i}}$ and utilizes only the sampled points for gradient computation ($|\cdot|$ denotes the set cardinality). This is to maintain the same complexity of the update step as FedAvg.\\
\textbf{DPMS. }
Following model training at round $t$, selected client $i$ forms updated encoder $w_{e,i}^{t}$ and classifier $w_{c,i}^{t}$; recall that the client received the global decoder $\mathbf{w_{d}}$ which is no longer being updated. If the client did not share latent representation information in the previous rounds, it utilizes $w_{e,i}^{t}$ to compute the means of latent representations for its $n$ most abundant classes; note that in this step we only utilize the original (real) data to form latent representations. Let us denote the computed means by $(\bar{z}_{i}^{\mathcal{C}_{i}^{1}},\dots, \bar{z}_{i}^{\mathcal{C}_{i}^{n}})$, where $\mathcal{C}_{i}^{1},\dots,\mathcal{C}_{i}^{n}$ denote the indices of the most abundant classes in the $i^{th}$ client's dataset. The means are then perturbed by an additive zero-mean Gaussian noise $ \mathcal{N}(0,\sigma^{2})$ (see Figure~\ref{dpms}). There are two benefits of adding the noise: first, it introduces diversity in the synthetically generated data; and, second, it endues the shared information with differential privacy (more on differential privacy in the next section). The noisy corruption of latent representation means, however, may be so severe to result in unusable synthesized data; we would like to identify if this is the case before the client communicates such means to the server. To this end, the client applies the global decoder $\mathbf{w_{d}}$ and utilizes the (highly accurate) local classifier $w_{c,i}^{t}$ to attempt recognizing the reconstructed data point. Specifically, the noisy latent representation means are formed as
\vspace{-0.05 in}
\begin{equation}
\label{addnoise}
    \tilde{z}_{i}^{\mathcal{C}_{i}^{j}} =  \bar{z}_{i}^{\mathcal{C}_{i}^{j}} + \delta,
\end{equation}
where $\delta \sim \mathcal{N}(0,\sigma^{2})$ and $j \in \{1, \dots, n\}$.
Using the global decoder, the client reconstructs $\tilde{x}^{\mathcal{C}_{i}^{j}}$ in the original space, and then applies the local encoder/classifier to find its prediction $\tilde{y}$,
\vspace{-0.05 in}
\begin{equation}
    \tilde{x}^{\mathcal{C}_{i}^{j}} = \mathbf{w_{d}}( \tilde{z}_{i}^{\mathcal{C}_{i}^{j}}),  \quad  \tilde{y} = w_{c,i}^{t}(w_{e,i}^{t}(\tilde{x}^{\mathcal{C}_{i}^{j}})).
\end{equation}
If the prediction of the classifier is correct, i.e., $\tilde{y} = \mathcal{C}_{i}^{j}$, we retain the noisy latent representation mean $\tilde{z}_{i}^{\mathcal{C}_{i}^{j}}$ used to synthesize the classified point; otherwise, we declare that  $\tilde{z}_{i}^{\mathcal{C}_{i}^{j}}$ is unusable and discard it. We continue this procedure until the number of usable noisy latent representation means in each abundant class $\mathcal{C}_{i}^{j}$ meets a predetermined quota  $\alpha$ (a tunable hyperparameter). Finally, each client sends its set of noisy encoded means $\mathcal{Z}_{i} = \{ (\tilde{z}_{i,1}^{\mathcal{C}_{i}^{1}},\dots,\tilde{z}_{i,\alpha}^{\mathcal{C}_{i}^{1}}),\dots,(\tilde{z}_{i,1}^{\mathcal{C}_{i}^{
n}},\dots,\tilde{z}_{i,\alpha}^{\mathcal{C}_{i}^{n}})\}$ and their corresponding labels to the server.\footnote{To ensure that discarding noisy latent representation means which fail the test does not affect privacy, we empirically evaluate the standard deviation of the noise present in the perturbed means that have passed the test and are consequently shared; in experiments, this standard deviation was verified to be virtually identical to $\sigma$ in (\ref{addnoise}), implying that sub-selecting the means has no impact on the level of privacy introduced by the noise corruption of latent representation means.} Following the DPMS step, the server appends $i$, $\mathcal{C}_{i}$ and $\mathcal{Z}_{i}$ to $\mathcal{A}$, $\mathcal{C}$ and $\mathcal{Z}$, respectively. In future training rounds, even if sampled again (which in large-scale systems is highly unlikely), client $i$ will not be asked to share latent representation information. Moreover, the client can now delete the global decoder $\mathbf{w_{d}}$ to free up memory; in fact, a selected client maintains the global decoder model for at most one round and thus its impact on average memory consumption is only minor. Further discussion of communication, computation and memory consumption can be found in supplementary material section 8.

It is worth pointing out that the assisting and benefiting clients each utilize their respective {\it local} 
encoder and {\it 
global} decoder: the former (in conjunction with its local classifier and global decoder) to learn latent data representations, the latter to synthesize artificial data. Such a strategy is desirable for two reasons. First, the local classifier achieves performance superior to the global classifier since the latter is agglomerated from local models that may have drifted apart; therefore, local classifiers are more trustworthy decision-makers for selecting usable noisy means. Second, we encode raw data using the local encoder of the assisting client but decode the latent representation means with the global decoder at the benefiting client; this helps separate the benefiting client's synthetic data from the assisting client's raw data in a way that is complementary to the separation induced by averaging latent information or adding noise. While its impact appears challenging to formalize analytically, such a strategy intuitively helps protect the assisting client's privacy. 

\vspace{-0.05 in}
\subsection{Privacy concerns}
\vspace{-0.05 in}
\label{diffential_privacy}
For completeness, we here define the differential privacy mechanism.
\newtheorem{definition}{Definition}
\vspace{-0.05 in}
\begin{definition}[Differential Privacy]
\label{dp}
A randomized mechanism $\mathcal{M}: \mathcal{D}\rightarrow \mathcal{R}$
satisfies $(\epsilon, \delta)$ differential privacy if for any two adjacent databases $d, d^{\prime} \in \mathcal{D}$ with only one different sample, and for any subset of the output $S \subseteq \mathcal{R}$, it holds that
\vspace{-0.1 in}
\begin{equation}
\operatorname{Pr}[\mathcal{M}(d) \in S] \leq e^{\epsilon} \operatorname{Pr}\left[\mathcal{M}\left(d^{\prime}\right) \in S\right]+\delta.
\end{equation}
The output of the random mechanism $\mathcal{M}$ is a random distribution; $\epsilon$ denotes an upper bound on the distance between distributions $\mathcal{M}(d)$ and $\mathcal{M}(d^{\prime})$ and can be interpreted as the privacy budget, while the relaxing factor $\delta$ is the probability that the $\varepsilon$-differential privacy is broken.
\end{definition}

\begin{definition}[Gaussian noise mechanism]
\label{Gaussian_mechanism}
The Gaussian noise mechanism achieving $(\varepsilon, \delta)$ differential privacy for any deterministic function $f: \mathcal{D} \rightarrow \mathbf{R}$ is defined as
\vspace{-0.05 in}
\begin{equation}
\mathcal{M}(d)=f(d)+\mathcal{N}\left(0, S_{f}^{2} \cdot \sigma^{2}\right),
\end{equation}
where $S_f$ denotes the maximum of the absolute distance $\left|f(d)-f\left(d^{\prime}\right)\right|$, and $\sigma> \sqrt{2 \log \frac{5}{4\delta}} / \varepsilon$. In other word, if we add a zero-mean Gaussian noise with variance $S_{f}^{2} \sigma^2$ to the output of $f$ and set the privacy budget $\epsilon$, the confidence of the resulting mechanism $\mathcal{M}$ is $\delta \geq \frac{4}{5}\exp{(-(\sigma\epsilon)^{2}/2)}$.
\end{definition}
While prior work on data augmentation in federated learning relied on data averaging for privacy protection \cite{feddg,fedmix}, differential attacks may be used to extract individual information from the averages. This motivates using concepts from differential privacy to quantify the privacy provided by noisy perturbations that FedDPMS injects into latent means. In related prior work, a number of methods that attempt to prevent privacy leaks from uploaded local gradients or models have been proposed in literature \cite{fredrikson2015model,hitaj2017deep,triastcyn2020federated, DP-CGAN}. These methods can readily be adapted to model training and client matching in FedDPMS. For simplicity, in this paper we limit our attention to characterizing differential privacy of sharing noisy latent means. To this end, consider $m$ latent representations $\mathbf{z_i} \in \mathcal{R}^{l}$, where $l$ is the dimension of latent representation, $1\leq i \leq m$; each client computes the $j^{th}$ element of the latent mean $\mathbf{\Bar{z}}$ by averaging the corresponding elements of the aforementioned $m$ vectors, $1\leq j \leq 
l$. Since clients reveal means of latent local data representations to the server, they may become exposed to the risk of leaking the latent representation of individual data points; this, in turn, could potentially be exploited to attempt reconstruction of raw data. To protect individual latent representations from a differential attack, we construct the following differential privacy mechanism. Let us interpret the averaging operation as a deterministic function $f(\mathbf{e})$, where vector $\mathbf{e}$ collects the $j^{th}$ elements of $m$ latent representations. Recall the definition of $S_{f}$,
\[S_f^{2} = \max \left( f(\mathbf{e}) - f(\mathbf{e}^{\prime})\right)^{2}, \]
where $\mathbf{e}^{\prime}$ is an adjacent input (similar to Def.~\ref{dp}) which matches $\mathbf{e}$ in $m-1$ elements. Since we use sigmoids as the activation functions of the layer generating latent representations (applied to all models in this paper), the latent means are between $[0,1]$; therefore,
\begin{equation*}
f(\mathbf{e}) - f(\mathbf{e}^{\prime}) \leq  \frac{\mathbf{1}^T\mathbf{e}^{\prime} + 1}{m} - \frac{\mathbf{1}^T\mathbf{e}^{\prime}}{m-1} 
 =  \frac{m-1 - \mathbf{1}^T\mathbf{e}^{\prime}}{m(m-1)} \leq \frac{1}{m},
\end{equation*}
where $\mathbf{1}^T$ denotes the vector of all ones having the same dimension as $\mathbf{e}^{\prime}$. Thus, $S_f = \frac{1}{m}$. For any $\epsilon > 0$ and $\delta > 0$, we can always identify $\sigma$ needed to ensure that $f(d)$ achieves $(\epsilon,\delta)$ differential privacy according to Definition~\ref{Gaussian_mechanism}. As an illustration, for $m=100$, $\epsilon = 0.5$ and $\delta = 0.01$, selecting $\sigma > 3.9$ (i.e., the standard deviation of noise $S_f\sigma > 0.039$), we are $99\%$ confident that the privacy is not broken. In fact, the standard deviation of noise added to latent means in our experiments is $>1$ while $m > 100$; thus there is abundant room for lower privacy budget $\epsilon$ or higher confidence $1-\delta$. In these settings, FedDPMS is very unlikely to break the privacy of the latent data representations.
\section{EXPERIMENTS}
\vspace{-0.05 in}
\label{exp}
\subsection{Datasets and Baselines}
\vspace{-0.05 in}
\label{exp setting}
We implemented all models and ran the experiments in Pytorch \cite{paszke2019pytorch}, using Adam \cite{kingma2014adam} optimizer with a learning rate 0.001 for all methods. The period of learning rate decay was set to 10, while the hyper-parameter $\gamma$ in Adam was set to $0.5$. We used the data batch size of $64$. Unless stated otherwise, the number of local epochs was set to $5$, while the number of global communication rounds was $50$; none of the methods improved performance by further increasing the number of global communication rounds. The default number of clients participating in federated learning is $10$. For simplicity and due to the relatively small number of clients, all clients participate in all rounds of the federated learning process.

To test the performance of our proposed framework, we conduct experiments on three datasets -- Fashion-MNIST \cite{xiao2017/online}, CIFAR10, and CIFAR100 \cite{CIFAR}. To control the degree of imbalance in the partitioned data, we utilize Dirichlet distribution \cite{fedmix,bayesian,moon} and generate non-iid partitions with varied concentration parameter $\beta$. Note that when the concentration parameter is very small, e.g. $\beta = 0.5$, a client may have very few samples (possible none) in some classes, potentially rendering the partition highly imbalanced. Examples of clients' local dataset class distributions are shown in supplementary material section 5. We compare the test accuracy of FedDPMS with four state-of-the-art federated learning methods including FedAvg \cite{mcmahan2017communication}, FedProx \cite{FedProx}, FedMix \cite{fedmix} and Moon \cite{moon}. The benchmarking experiments utilize VAEs and a convolutional neural network trained to perform image classification tasks. For detailed specification of the network architecture and hyper-parameters, please see  supplementary material section 2. 
\vspace{-0.05 in}
\subsection{Results Analysis}
\vspace{-0.05 in}
\noindent \textbf{Accuracy comparison.}
 \begin{figure*}[h] 

    \centering
	  \subfloat[FMNIST]{
       \includegraphics[width=0.32\linewidth]{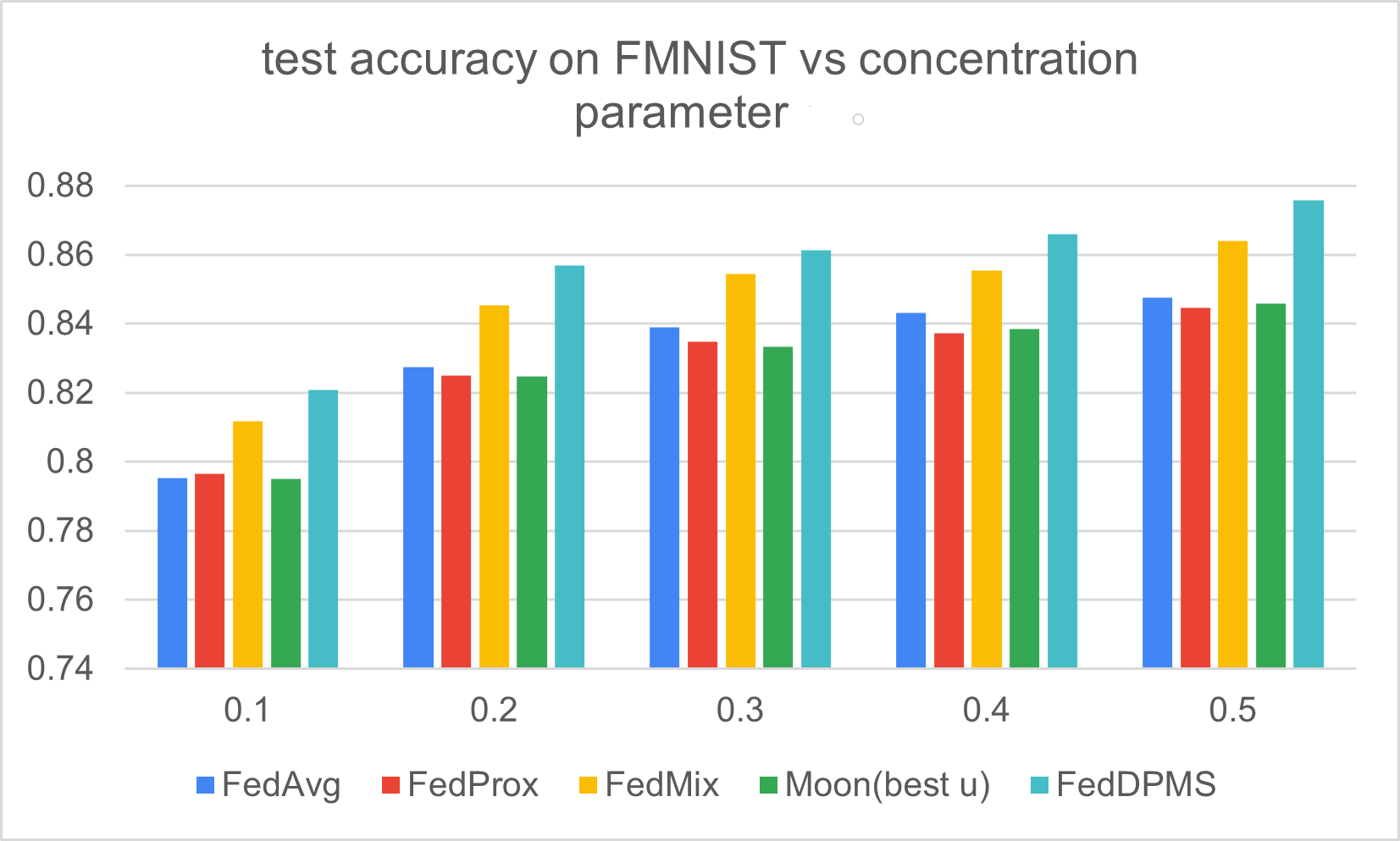}}
    \label{1a}
	  \subfloat[CIFAR10]{
        \includegraphics[width=0.32\linewidth]{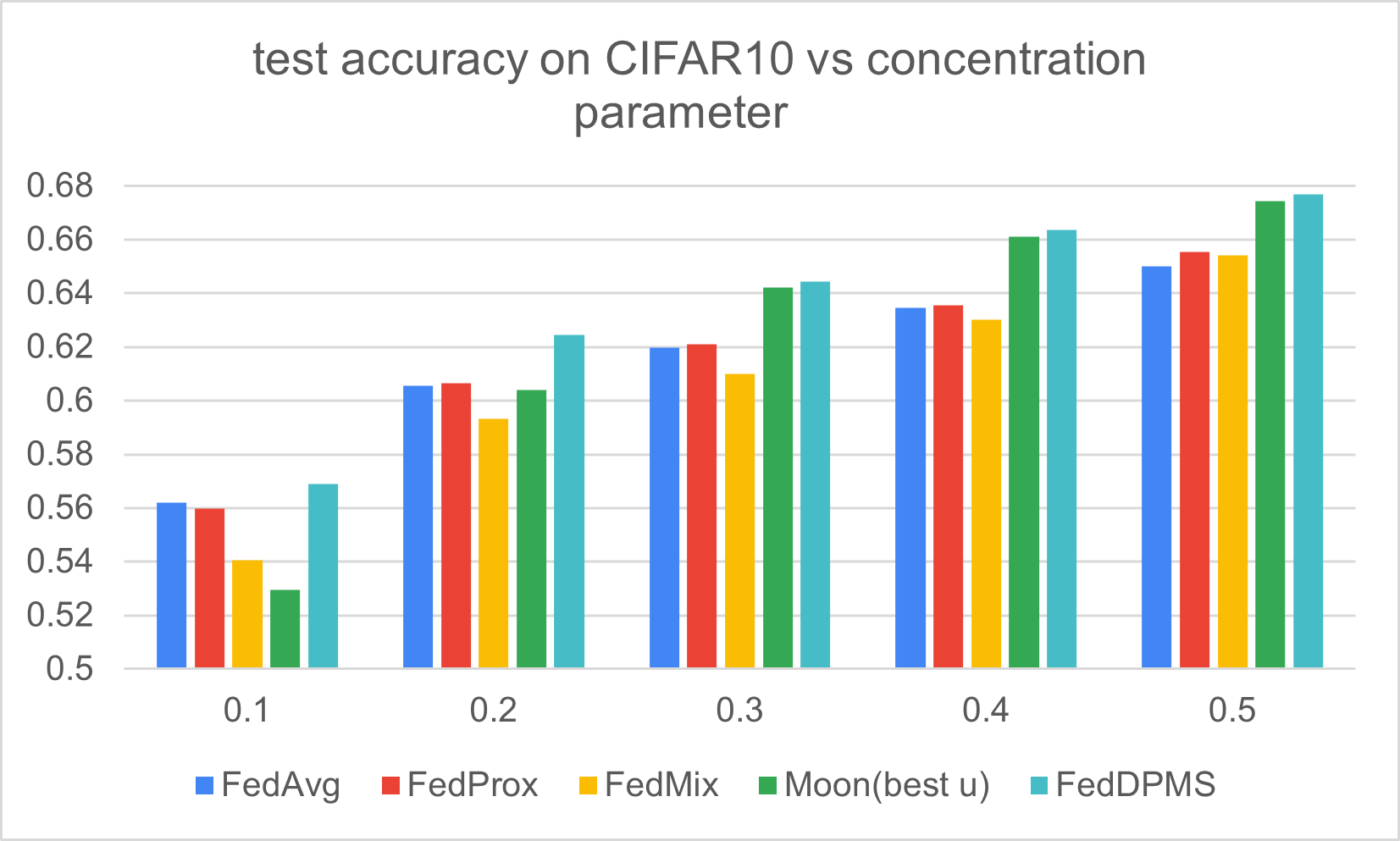}}
    \label{1b}
	  \subfloat[CIFAR100]{
        \includegraphics[width=0.32\linewidth]{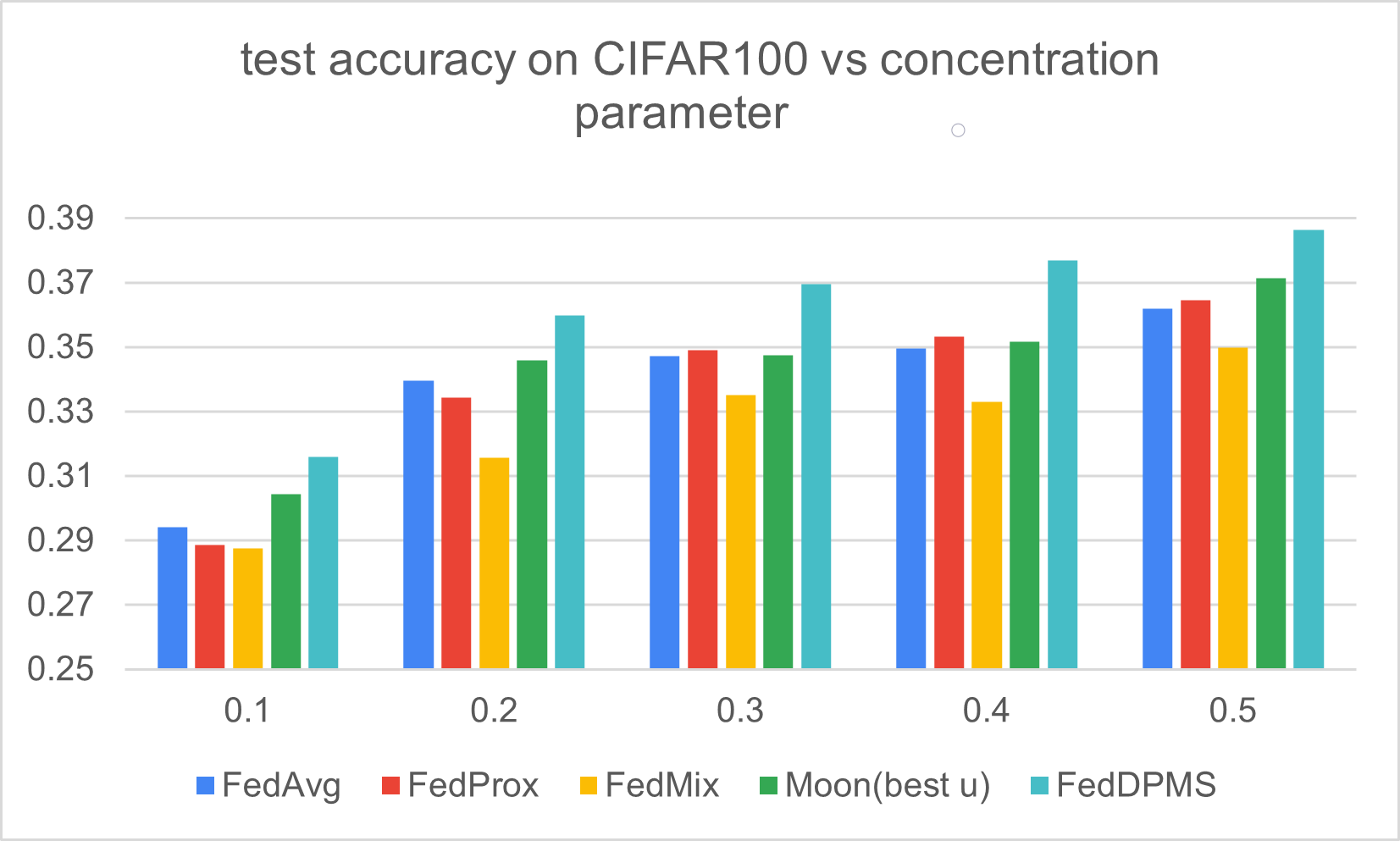}}
    \label{1c}\hfill
	  
	  \caption{Test accuracy of different approaches as the concentration parameter $\beta$ takes values from $\{0.1,0.2,0.3,0.4,0.5\}$. We run five trials with different random seeds and report the mean accuracy.}
	  \label{differentbeta} 
\end{figure*}
 \begin{figure*}[h] 
    \centering
	  \subfloat[FMNIST]{
       \includegraphics[width=0.32\linewidth]{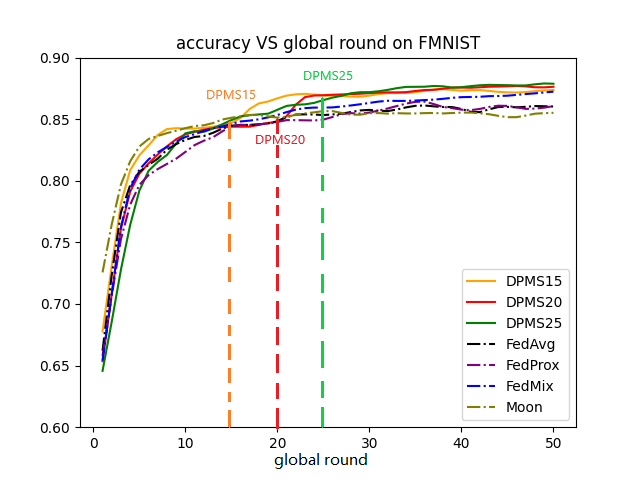}}
    \label{1a}
	  \subfloat[CIFAR10]{
        \includegraphics[width=0.32\linewidth]{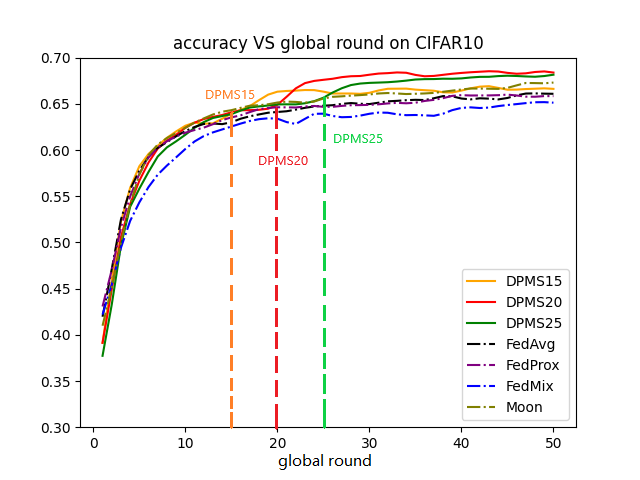}}
    \label{1b}
	  \subfloat[CIFAR100]{
        \includegraphics[width=0.32\linewidth]{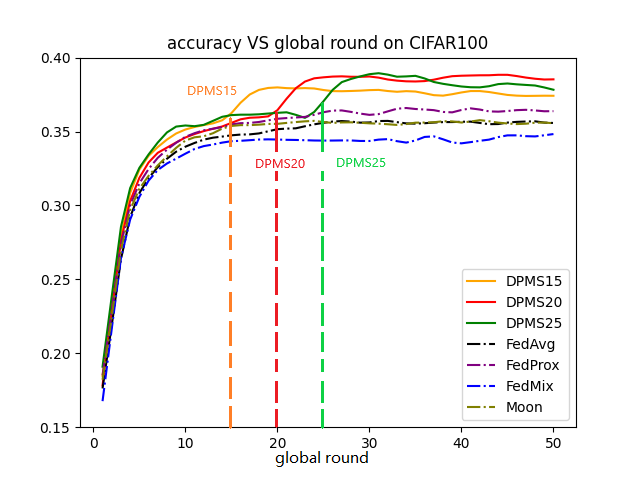}}
    \label{1c}\hfill
	  
	  \caption{Test accuracy of different approaches on FMNIST, CIFAR10 and CIFAR100. All experiments are conducted with the concentration parameter $\beta = 0.5$ and 10 clients. We run three trials with different random seeds and report the mean accuracy.}
\label{convergence} 
\vspace{-0.15 in}
\end{figure*}
 Table \ref{table1} compares the test accuracy of FedDPMS with the concentration parameter $\beta =  0.5$ and default parameter settings against the competing methods. In such a severely heterogeneous setting, relative improvement of FedDPMS over FedAvg is $3.3\%$, $4.5\%$ and $6.6\%$ on FMNIST, CIFAR10 and CIFAR100, respectively; overall, FedDPMS achieves the best performance among all approaches. As for the performance of other methods relative to each other: FedMix has a significant advantage over FedAvg on FMNIST while closely tracking, along with FedProx, performance of FedAvg on the other two datasets. Moon exhibits a slight improvement over FedAvg on CIFAR100 and achieves close 2nd (behind FedDPMS) performance on CIFAR10.
 \begin{table}[h]
\caption{Test accuracy of FedDPMS and the competing methods on FMNIST, CIFAR10 and CIFAR100 datasets. We run five trials with different random seeds and report the mean accuracy.}
\vspace{-0.1 in}
\label{table1}
\begin{center}
\begin{small}
\begin{sc}
\begin{tabular}{lcccr}
\hline
 Scheme & FMNIST & CIFAR10 & CIFAR100 \\
\hline
FedAvg    & 0.8476 & 0.6501& 0.3621 \\
FedProx     & 0.8446& 0.6553& 0.3646\\
FedMix & 0.8640  & 0.6542 & 0.3498
   \\
Moon    & 0.8458& 0.6742&  0.3715 \\
FedDPMS    & \textbf{0.8759}& \textbf{0.6797}& \textbf{0.3861} \\
\hline
\end{tabular}
\end{sc}
\end{small}
\end{center}
\vspace{-0.25 in}
\end{table}
\\ \textbf{Effect of data heterogeneity.}
As previously stated, the concentration parameter $\beta$ controls heterogeneity of the data partitions -- smaller $\beta$ leads to more severe class imbalance. We use the Dirichlet distribution with $\beta = \{0.1,0.2,0.3,0.4,0.5\}$ to generate varying data partitions and study the effect of heterogeneity on different methods. As the results in Fig. \ref{differentbeta} demonstrate, FedDPMS consistently achieves the best performance on FMNIST and CIFAR10/100. Among other methods, FedProx closely tracks FedAvg in all experiments, while Moon is competitive on CIFAR10 yet performs badly for $\beta = 0.1$. The takeaway from competitive performance of FedMix on FMNIST, but poor on both CIFAR10 and CIFAR100, is that utilizing interpolation to synthesize samples from raw data succeeds on simple datasets but does not on more complex ones; moreover, the approach deployed by FedMix is risky in terms of potential privacy leaks. In all, the experiments demonstrate consistently best performance of FedDPMS across different data distributions and levels of heterogeneity.
\\ \textbf{Scalability.}
So far, our experiments involved simulating federated learning systems with 10 clients. To investigate the scalability of FedDPMS, we vary the number of clients in experiments on CIFAR100. In particular, the number of clients is varied across $\{10,20,30,40,50\}$; to maintain the same amount of data per client as the number of clients grow, an increasing fraction of CIFAR100 data ($10\%,20\%, 30\%, 40\%, 50\%$) is partitioned and allocated to the clients in these experiments. Meanwhile, we keep the default heterogeneity setting $\beta=0.5$. The results of the experiments are reported in Table \ref{table2}, showing that FedDPMS outperforms all other approaches. For example, relative improvement of FedDPMS over FedAvg is more than $22.3\%, 15.6\%$, $7.3\%$, $7.7\%$ and $4.6\%$ in the experiments involving $10$-$50$ clients, respectively. 
\begin{table}[t]
\caption{Test accuracy as the number of clients varies from $10$ to $50$ in steps of $10$; the clients utilize $10\%$-$50\%$ samples of CIFAR100 with class partitions generated via Dirichlet distribution with the concentration parameter $\beta = 0.5$. We run three trials with different random seeds and report the mean accuracy.} 
\vspace{-0.1 in}
\label{table2}
\begin{center}
\begin{small}
\begin{sc}
\begin{tabular}{lp{0.8cm}p{0.8cm}p{0.8cm}p{0.8cm}p{0.8cm}}
\hline
SCHEME & 10 & 20 & 30 & 40 & 50 \\
\hline  
FedAvg     & 0.1881& 0.2401& 0.2738 &0.2842 &0.3066 \\
FedProx     &0.1844 & 0.2408 & 0.2780 &0.2914 &0.3081\\
FedMix     & 0.1820 &0.2365& 0.2676 &0.2877& 0.3045  \\
Moon    &0.1871 & 0.2375&  0.2707 &0.2912 &0.3095 \\
FedDPMS   & \textbf{0.2301} & \textbf{0.2695}& \textbf{0.2940} &\textbf{0.3061} &\textbf{0.3208}\\
\hline
\end{tabular}
\end{sc}
\end{small}
\end{center}
\vspace{-0.25 in}
\end{table}
\\ \textbf{Convergence rate.} Fig. \ref{convergence} shows the test accuracy of all approaches across the federated training rounds. For FedDPMS, we further experimented by varying duration of the preliminary training phase ($15$, $20$ and $25$). Among all methods, FedDPMS exhibits the fastest convergence rate and best accuracy on CIFAR10 and CIFAR100. FedMix has the slowest convergence rate on CIFAR10/100 but a slightly higher accuracy than FedDPMS on FMNIST. After sharing the latent representation means and augmenting local datasets with synthetic data, the test accuracy of each FedDPMS model suddenly increases. The results imply that the quality of augmented data improves with the number of preliminary training rounds; however, the longer we wait, the smaller the number of rounds to train the model on the augmented dataset. The experiments suggest that setting the number of preliminary training rounds to $40\%$ of the total number of allotted rounds is a suitable choice.
\vspace{-0.05 in}
\section{Conclusion}
\vspace{-0.05 in}
Data heterogeneity is a critical challenge hindering practical federated learning systems, potentially causing major performance deterioration. We propose a novel framework, FedDPMS, aiming to enable accurate and robust performance of federated deep learning models trained on heterogeneous distributed datasets. This is accomplished by sharing differentially-private information which the clients use to enrich local datasets and thus combat the local model drift. As our experimental results demonstrate, FedDPMS outperforms state-of-the-art federated learning methods on image classification tasks with varied levels of heterogeneity across clients while requiring only a minor increase in communication cost. FedDPMS does requires additional computation and memory resources, the amount of which depends on the specifics of the training process. In applications where the high accuracy is imperative, FedDPMS provides an attractive and effective framework for overcoming challenges presented by data heterogeneity.
{\small
\bibliographystyle{ieee_fullname}
\bibliography{egbib}
}

\end{document}